\documentclass[letterpaper, 10 pt, conference]{ieeeconf}  

\IEEEoverridecommandlockouts                              

\usepackage{graphics} 
\usepackage{epsfig} 
\usepackage{amsmath} 
\usepackage{amssymb}  
\usepackage{color,soul} 
\usepackage{bm} 
\usepackage{flushend}
\usepackage[table]{xcolor}
\usepackage{multicol}

\renewcommand{\matrix}{\boldsymbol}
\newcommand*{\rev}{\textcolor{black}}

\title{\LARGE \bf
Optimal control of differentially flat underactuated planar robots in the perspective of oscillation mitigation
}

\author{Stefano Lovato$^{1}$, Michele Tonan$^{1}$, Matteo Bottin$^{1}$, Matteo Massaro$^{1}$, Alberto Doria$^{1}$, Giulio Rosati$^{1}$
\thanks{Project financially supported by BIRD 2023 Research Program of University of Padova grant number BOTT\_BIRD23\_01.}
\thanks{$^{1}$All the authors are with the Department of Industrial Engineering, University of Padova, Via Venezia 1, Padova, Italy
{\tt\small matteo.bottin@unipd.it}}
}

\begin{document}

\maketitle
\thispagestyle{empty}
\pagestyle{empty}

\begin{abstract}

Underactuated robots are characterized by a larger number of degrees of freedom than actuators and, if they are designed with a specific mass distribution, they can be controlled by means of differential flatness theory. This \rev{dynamic} property enables the development of lightweight and cost-effective robotic systems with enhanced dexterity. However, a key challenge lies in managing the elastic-passive joints, whose control demands precise and comprehensive dynamic modeling of the system. 
To simplify dynamic models, particularly for low-speed trajectories, friction is often neglected. While this assumption simplifies analysis and control design, it introduces residual oscillations of the end-effector about the target position. 
In this paper, the possibility of using optimal control along with differential flatness control is investigated to improve the tracking of the planned trajectories. First, the study was carried out through formal analysis, and then, it was validated by means of numerical simulations.

Results highlight that optimal control can be used to plan the flat variables considering different (quadratic) performance indices: control effort, i.e. motor torque, and potential energy of the considered underactuated joint. Moreover, the minimization of potential energy can be used to design motion laws that are robust against variation of the stiffness and damping of the underactuated joint, thus reducing oscillations in the case of stiffness/damping mismatch.


\end{abstract}

\section{Introduction}

\rev{Underactuated planar robots}
 are dynamic systems with some joints without direct actuation, and the \rev{corresponding} links are connected via elastic elements to the rest of kinematic chain. In such systems, actuated joints indirectly control the motion of the elastic joints.

Research in this area has expanded, leading to applications such as jointed arm robots \cite{Agrawal2006,Firouzeh20152536,qin2022design}, cable-driven robots \cite{Barbazza2017896,Zanotto20113964}, and walking robots \cite{Gupta2017607,He2019}. These systems are attractive for their reduced weight and cost, and increased dexterity with fewer actuators \cite{Bottin2019293,Tonan2023}. Other notable examples include prosthetic devices \cite{KelvinLoutanJr2023}, sea surface robots \cite{MicheleAngelini2023}, and advanced grippers \cite{Becedas2011,BoiOkken2023,ma2016spherical}.
However, underactuated systems are hard to control, and over the years, many solutions have been proposed to overcome their limitations  \cite{RichiedeiJMR,XIN2023111280}.
Differential flatness, a property allowing system states to be expressed via flat outputs thanks to specific inertial properties \cite{sira2018differentially,Murray1995349}, has facilitated control strategies for both linear and nonlinear systems \cite{Yong20153898,Zanotto2013,Ryu20105201,Mounier2008445}. For underactuated robots, flatness-based control has been validated in simulations and experiments, both for point-to-point and more complex motions \cite{Franch2010548,Franch2013161,Sangwan20082423,Bottin2022}. Over the years, disturbances, e.g. friction, have been included in the mathematical model \cite{Tonan20241262}. 

Robot trajectories, whether for fully actuated or underactuated systems, can be optimized with respect to various performance criteria, such as minimizing motion time \cite{Lovato2023} or reducing energy consumption \cite{506579}. 
Within the framework of differentially flat systems, researchers showed that optimal control methods can be adopted to plan feasible trajectories \cite{Logan2024}.
Further comparative studies evaluated the effectiveness of Linear Quadratic Regulator (LQR) control against differentially flat control strategies, particularly in applications involving linear guide feedback systems \cite{gomez2015optimal}.




This paper focuses on how optimal control can be used in the trajectory planning of underactuated differentially flat planar robots to achieve better performance in terms of control effort, i.e. motor torque, or oscillation mitigation at the end of the motion.
The paper is structured as follows.
Section \ref{sec:math} presents the mathematical model of serial differential flat robots.
Section \ref{sec:optimal} describes the optimal control planning strategies.
Section \ref{sec:numerical} shows some simulations and describes the results.
Section \ref{sec:conclusions} draws the conclusions.

\section{Mathematical model}\label{sec:math}
\begin{figure}[t!]
    \centering
    \includegraphics[width=1\linewidth]{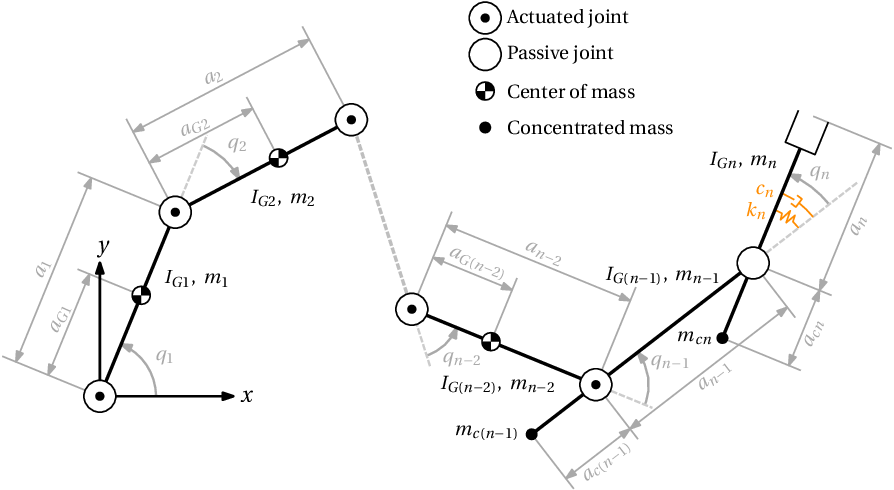}
    \caption{Scheme of the underactuated differentially flat robot with $n$-DOF.}
    \label{fig:scheme_system_nDOF}
\end{figure}
The robot considered in this paper presents one elastic-passive joint and is differentially flat since it is designed according to \cite{Franch2010548}. 
In particular, a $n$-DOF robot with only one elastic-passive joint is differentially flat if the center of mass of the final link ($n$) is on the $n$-th joint axis and if the center of mass of both links $n$ and $n-1$ is on the $(n-1)$-th joint axis.
The actuators for the system are located on all joints but the last, which is equipped with a torsional spring of stiffness $k_n$ and a viscous damper with damping coefficient $c_n$.

Figure \ref{fig:scheme_system_nDOF} shows the scheme of an $n$-DOF robot with the last elastic-passive joint. $q_1$ represents the rotation of link 1 with respect to the base frame $x-y$, and $q_i$ is the relative rotation of link $i$ with respect to $i-1$ link.
Mass and barycentric moment of inertia of the $i$-th link are $m_i$ and $I_{Gi}$, respectively; the distance of the center of mass of $i$-th link from the $i$-th joint and the link total length are $a_{Gi}$ and $a_{i}$ respectively (please note that $a_{G(n-1)}=a_{Gn}=0$ since the two links are fully balanced).
The last two fully balanced links are equipped with two counterbalancing masses, namely $m_{c(n-1)}$ and $m_{c(n)}$, placed at a distance $a_{C(n-1)}$ and $a_{C(n)}$ from the respective joint. 

The dynamic model of the system can be derived by means of the Lagrangian approach in matrix form:
\begin{equation}
    \matrix{M}_n(\matrix{q})\matrix{\ddot q}
    + \matrix{C}_n\matrix{\dot q}
    + \matrix{K}_n \matrix{q} +
     \matrix{b}(\matrix{q},\matrix{\dot{q}}) + \matrix{g}(\matrix{q})
    = \matrix{\tau}
    \label{Eq:sist_eq_nDOF_matrix}
\end{equation}
where \rev{vectors} $\matrix{q}$, $\matrix{\dot q}$ and $\matrix{\ddot q}$ contain all joint variables and their first and second derivatives, \rev{vector} $\matrix{b}(\matrix{q},\matrix{\dot{q}})$ contains all Coriolis-centrifugal terms and \rev{vector} $\matrix{g}(\matrix{q})$ contains all gravitational terms.
Please note that the last two terms of both $\matrix{b}(\matrix{q},\matrix{\dot{q}})$ and $\matrix{g}(\matrix{q})$ are null since the links are fully balanced. $\matrix{M}_n$, $\matrix{C}_n$ and $\matrix{K}_n$ are the mass, damping, and stiffness matrices, defined as:
\begin{equation}
\begin{small}
\begin{gathered}
\label{eq:MnCnKn}
         \matrix{M}_n (\matrix{q}) = \left[ \begin{array}{ *{5}{c} }
                    & & &  I_{n-1}^*  & I_{n}^* \\
                    & & & I_{n-1}^*  & I_{n}^*  \\
                    & & & \vdots & \vdots\\
                    \multicolumn{3}{c}{\raisebox{\dimexpr\normalbaselineskip+.7\ht\strutbox-.5\height}[0pt][0pt]
                        {{$\matrix{I}^*_{(n-2)\times (n-2)}(\matrix{q})$}}} & I_{n-1}^*  & I_{n}^* \\
                     I_{n-1}^* & \cdots & I_{n-1}^* & I_{n-1}^*  & I_{n}^*\\
                     I_{n}^* & \cdots & I_{n}^* & I_{n}^*  & I_{n}^*\\
                  \end{array} \right]\\
        \matrix{C}_n = \left[ \begin{array}{ *{4}{c} }
                    & & & 0 \\
                    & & & \vdots  \\
                    \multicolumn{3}{c}{\raisebox{\dimexpr\normalbaselineskip+.7\ht\strutbox-.5\height}[0pt][0pt]
                        {{$\matrix{0}_{(n-1)\times (n-1)}$}}} & 0 \\
                    0 & \cdots & 0 & c_n
                  \end{array} \right], \
        \matrix{K}_n = \left[ \begin{array}{ *{4}{c} }
                    & & & 0 \\
                    & & & \vdots  \\
                    \multicolumn{3}{c}{\raisebox{\dimexpr\normalbaselineskip+.7\ht\strutbox-.5\height}[0pt][0pt]
                        {{$\matrix{0}_{(n-1)\times (n-1)}$}}} & 0 \\
                    0 & \cdots & 0 & k_n
                 \end{array} \right] 
\end{gathered}
\end{small}
\end{equation}
5
%
%
%
It is worth noting that the last two rows of matrix $\matrix{M}_n$ are constant, i.e., are independent for any robot configuration $\matrix{q}$. 
\rev{As explained in \cite{Agrawal2006}}, from \rev{Eqs.\eqref{Eq:sist_eq_nDOF_matrix} and \eqref{eq:MnCnKn}} differential flatness can be used to define a set of flat variables $\matrix{y}(t)$ so that:
\begin{equation}\label{eq:y}
    y_1=\sum_{i=1}^nq_i\hspace{0.5cm},\hspace{0.5cm}
    y_k=q_{k-1}\ \text{with}\ k=2,\cdots,n-1
\end{equation}
where $y_1$ has the physical meaning of absolute orientation of the last link with respect to the fixed reference frame.

Exploiting the flat variables, the motion of the last joint can be controlled \cite{Tonan2024}:
\begin{equation}\label{eq:qn}\begin{gathered}
    q_n = -\frac{I_n^*}{k_n}\ddot{y}_1 + \frac{I_n^*c_n}{k_n^2}y_1^{(3)}, \quad
    \ddot q_n = -\frac{I_n^*}{k_n}y_1^{(4)} + \frac{I_n^*c_n}{k_n^2}y_1^{(5)}
\end{gathered}
\end{equation}
where $y_1^{(i)}=\frac{d^i y_1}{dt^i}$ is the $i$-th time derivative of $y_1$.
%
\rev{If acceleration $\ddot q_{n}$ of Eq.\eqref{eq:qn} and the sum of the other accelerations expressed as a function of flat variables $\matrix{y}(t)$ are introduced in the $n-1$ equation of the motion,} motor torque $\tau_{n-1}$ of joint $n-1$ can be calculated \rev{in the flat variable space} \cite{Tonan20241262}:
\rev{\begin{equation}\label{eq:tau}
    \tau_{n-1} = I_{n-1}^* \ddot y_1 + \frac{I_n^*\left(I_{n-1}^* - I_n^*\right)}{k_n}  \left( y_1^{(4)} - \frac{c_n}{k_n} y_1^{(5)}\right) 
\end{equation}}
%
%
The control of $q_n$ can be achieved by a proper planning of the flat variable $y_1$.
In particular, it is important to notice that the dynamic model contains the acceleration (Eq.\eqref{Eq:sist_eq_nDOF_matrix}), hence the flat variable $y_1$ must be a $C^5$ function, and 6 conditions must be imposed on both the initial and final configurations, for a total of 12 boundary conditions.
Such conditions are the initial and final joint angles, as well as the null derivatives.
Additional constraints can be imposed if there is the need for passing through via points \cite{Tonan2024}.

It is worth noting that Eq.\eqref{eq:tau} is achieved by employing a Taylor expansion of \rev{the viscous damping term that holds true for $\frac{c_n}{k_n}\omega\ll1$ (where $\omega$ is the angular frequency) and implies a small approximation. Therefore}, the analytical result is not exact, and some oscillations may appear which result in small errors at the end of the motion \cite{Tonan20241262}.


\section{Optimal motion planning}\label{sec:optimal}

Previous work \cite{Tonan2024} has shown that in order to enforce the 12 boundary conditions, an 11th degree polynomial is sufficient to ensure a $C^5$ function for the flat variable $y_1$. 
Nevertheless, adopting a different motion law for $y_1$ can be advantageous when the objective is to optimize other factors in the trajectory, such as the control effort. 

In this section, a more general framework for motion planning is presented, which is based upon the Euler-Lagrange equation (ELE) \cite{troutman2012variational} applied to a quadratic performance index. The cost function $\mathcal J$ is assumed to be of the form
\begin{equation}\label{eq:cost}
    \mathcal J[y_1(t)] = \int_{0}^{T} \mathcal{L}(y_1^{(1)},...,y_1^{(6)}) dt = \int_0^T \frac{1}{2}(\bm y_1)^\text{T} \bm Q (\bm y_1) dt
\end{equation}
where \rev{$T$ is the motion time}, $\mathcal{L}$ is the integral cost, $\bm y_1=[y_1^{(1)},y_1^{(2)},...,y_1^{(6)}]^T$ is a vector containing the flat variable derivatives from order 1 to order 6, and \rev{$\bm Q = [Q_{ij}]$} is a $6\times6$ symmetric positive-semidefinite weighting matrix\rev{; without loss of generality, $Q_{66}=1$ is enforced}. The ELE for an integral cost including derivatives up to 6th order is
\begin{equation}
    \sum_{k=0}^{6} (-1)^k \frac{d^k}{dt^k} \left(\frac{\partial \mathcal{L}}{\partial y_1^{(k)}}\right) = 0
\end{equation}
which applied to Eq.\eqref{eq:cost} gives the following 12-order constant-coefficient linear ordinary differential equation
\begin{eqnarray}
    y_1^{(12)} - (Q_{55}-2Q_{64}) y_1^{(10)} +  \nonumber \\
    (Q_{44}-2Q_{53}+2Q_{62}) y_1^{(8)}+ \nonumber\\ 
    -(Q_{33}+2Q_{51}-2Q_{42})y_1^{(6)} + \label{eq:ode}  \\
    (Q_{22}-2Q_{31}) y_1^{(4)} - 
    Q_{11} y_1^{(2)}  = 0  \nonumber
\end{eqnarray}
%
The solution of Eq.\eqref{eq:ode} is obtained from the roots of the associated characteristic polynomial. The general solution is expressed as a linear combination of exponential and oscillatory terms---depending on the root complexity---and involves 12 coefficients, determined by enforcing the 12 boundary conditions.
%
\rev{
The solution of Eq.\eqref{eq:ode} is obtained from the roots of the associated characteristic polynomial. Real roots $\beta$ of multiplicity $m$ contribute with terms of the form
\begin{equation}
    e^{\beta t} \sum_{k=0}^{m-1} C_k t^k,
\end{equation}
while pairs of complex conjugate roots $\alpha \pm i\omega$ with multiplicity $m$ yield
\begin{equation}
    e^{\alpha t} \left[
    \left(\sum_{k=0}^{m-1} A_k t^k\right)\cos(\omega t) +
    \left(\sum_{k=0}^{m-1} B_k t^k\right)\sin(\omega t)
    \right].
\end{equation}
The general solution involves 12 coefficients, determined by enforcing the 12 boundary conditions.
}


Depending on the selected performance index, different motion laws $y_1(t)$ are obtained.
In this work, three performance indexes are separately considered. 

\subsection{Polynomial planning}
The first performance index aims to minimize the \rev{root mean square (RMS)} value of $y_1^{(6)}$, \rev{and it actually gives} give the polynomial planning used in \cite{Tonan2024}. The integral cost $\mathcal{L}$ is
\begin{equation}
    \mathcal{L} = (y_1^{(6)})^2
\end{equation}
which means $Q_{66}=1$ and the other coefficients of $\bm Q$ are null. In such a case, Eq.\eqref{eq:ode} reduces to
\begin{equation}
    y_1^{(12)} = 0
\end{equation}
whose solution is a 11-th degree polynomial. 
\rev{As a remark, it is worth noting that high-order polynomials can be poorly conditioned, and the use of alternative bases may improve numerical robustness. In this work, appropriate scaling of the system is employed to mitigate this effect.}

\subsection{Min-control-effort planning}
The second performance index \rev{here} considered aims to minimize \rev{a combination of the RMS value of $y_1^{(6)}$ along with} the control effort, represented by motor torque $\tau_{n-1}$. Therefore, in this case, the integral cost $\mathcal{L}$ is given by
\begin{equation}
    \mathcal{L} = (y_1^{(6)})^2 + r^8 \left[\tau_{n-1}\left(\ddot y_1, (y_1^{(4)}), (y_1^{(5)}) \right)\right]^2 \label{eq:mintau}
\end{equation}
where $r$ is weighting factor used to tune the minimization\footnote{\label{note1}It is worth noting that $r$ and $p$ are elevated to the power of 8 so that the values of $r$ and $p$ can be chosen below 1000. In other words, the power of 8 acts only as a scaling factor, and has no physical meaning.}. The coefficients in Eq.\eqref{eq:ode} are
\begin{equation}\label{eq:Min-control-effort}
\begin{small}\begin{aligned}
        Q_{22}= r^8 I^{*2}_{n-1}, &\
    Q_{42}= r^8 \frac{I^*_{n-1}I^*_{n}(I^*_{n-1}-I^*_{n})}{k_n} \\
    Q_{44}= r^8\frac{I^{*2}_{n}(I^*_{n-1}-I^*_{n})^2}{k_n^2}, &\
    Q_{52}= r^8 \frac{c_n I^*_{n-1}I^*_{n}(I^*_{n-1}-I^*_{n})}{k_n^2} \\
    Q_{54}= r^8\frac{c_n I^{*2}_{n}(I^*_{n-1}-I^*_{n})^2}{k_n^3}, &\
    Q_{55}= r^8\frac{c_n^2 I^{*2}_{n}(I^*_{n-1}-I^*_{n})^2}{k_n^4}
\end{aligned}\end{small}
\end{equation}
It is worth noting that when $r\to\infty$, Eq.\eqref{eq:ode} becomes a 10-order ordinary differential equation, with only 10 coefficients being determined by the boundary conditions. In such case, not all the specified boundary conditions can be arbitrarily enforced. 

\subsection{Min-potential-energy planning}
The third performance index \rev{here} introduced aims to minimize \rev{a combination of the \rev{RMS} value of $y_1^{(6)}$ along with} the potential energy $U_n=\frac{1}{2}k_n q_n^2$ accumulated in the underactuated joint. This performance index is motivated by the desire of reducing the influence of the joint stiffness and damping (i.e. $k_n,c_n$) on the actual motion of the system.
Using the flat variables, the motion of the last joint $q_n$ is expressed as a function of $\ddot y, y_1^{(3)}$; see Eq.\eqref{eq:qn}. Under the assumption of small $c_n$, the dominant contribution \rev{to $q_n$} is given by $\ddot y_1$. Consequently, to minimize the potential energy $U_n$, the selected integral cost is a combination of $(y_1^{(6)})^2$ and $(\ddot y_1)^2 \propto U_n $ and is defined as
\begin{equation}
    \mathcal{L} = (y_1^{(6)})^2 + p^8 \ddot y_1 ^2 \label{eq:minpot}
\end{equation}
where $p$ is weighting factor used to tune the minimization\footnote{See footnote \ref{note1}}.
It yields:
\begin{equation}\label{eq:odeminpot}
    y_1^{(12)} - p^8 y_1^{(4)} = 0
\end{equation}
A closed-form solution exists in this case and is
\begin{equation}
\begin{small}
\begin{gathered}
    y_1(t) = C_1 + C_2 t + C_3 t^2 + C_4 t^3  \\
    + e^{\frac{\sqrt{2 + \sqrt{2}}}{2}pt} 
\left[ A_1 \cos\left( \frac{\sqrt{2 + \sqrt{2}}}{2}pt \right) + B_1 \sin\left( \frac{\sqrt{2 + \sqrt{2}}}{2}pt \right) \right] \\
    + e^{\frac{\sqrt{2 - \sqrt{2}}}{2}pt} 
\left[ A_2 \cos\left( \frac{\sqrt{2 + \sqrt{2}}}{2}pt \right) + B_2 \sin\left( \frac{\sqrt{2 + \sqrt{2}}}{2}pt \right) \right] \\
    + e^{\frac{\sqrt{2 + \sqrt{2}}}{2}pt} 
\left[ A_3 \cos\left( \frac{\sqrt{2 - \sqrt{2}}}{2}pt \right) + B_3 \sin\left( \frac{\sqrt{2 - \sqrt{2}}}{2}pt \right) \right] \\
    + e^{\frac{\sqrt{2 - \sqrt{2}}}{2}pt} 
\left[ A_4 \cos\left( \frac{\sqrt{2 - \sqrt{2}}}{2}pt \right) + B_4 \sin\left( \frac{\sqrt{2 - \sqrt{2}}}{2}pt \right) \right] 
\end{gathered}
\end{small}
\end{equation}
with the 12 coefficients $C_1,...C_4,A_1,...,A_4,B_1,...,B_4$ determined by the boundary conditions.
When $p\to\infty$, Eq.\eqref{eq:odeminpot} becomes $y_1^{(4)} = 0$, whose solution is a 3rd degree polynomial, with only 4 coefficients being determined by the boundary conditions. Again, not all the specified boundary conditions can be arbitrarily enforced in this case. 

As a final remark, motion planning with mixed-minimization can also be considered, i.e. Eq.\eqref{eq:mintau} together with \eqref{eq:minpot}. The corresponding integral cost is
\begin{equation}
    \mathcal{L} = (y_1^{(6)})^2 + r^8 \left[\tau_{n-1}\left(\ddot y_1, (y_1^{(4)}), (y_1^{(5)}) \right)\right]^2 + p^8 \ddot y_1 ^2
\end{equation}
which gives the same coefficients of Eq.\eqref{eq:Min-control-effort} with the only exception of $Q_{22}$ which is:
\begin{equation}
    Q_{22} = r^8 I^{*2}_{n-1} + p^8
\end{equation}

\section{Example of application on a 2-DOF underactuated robot}\label{sec:numerical}
\rev{Since flatness‑based control has already been successfully implemented and validated on similar real-world robots, even taking into account friction effects \cite{Sangwan20082423,Tonan20241262,tonan2026design}, the present study evaluates the proposed control strategy through numerical experiments. Hence,}
the proposed optimal motion planning is applied to a planar 2-DOF robot with one motor and one underactuated joint. The robot consists of two links, with the second one fully balanced. The joint variables are $\matrix{q}=[q_1, q_2]^T$, the matrices in Eq.\eqref{eq:MnCnKn} are
\begin{equation}\label{eq:M2C2K2}
\begin{gathered}
    \matrix{M}_n = \left[ \begin{array}{ *{2}{c} }
    I_1^* & I_2^* \\ I_2^* & I_2^*
    \end{array} \right], \ 
    \matrix{C}_n = \left[ \begin{array}{ *{2}{c} }
    0 & 0 \\0 & c_2
    \end{array} \right], \
    \matrix{K}_n = \left[ \begin{array}{ *{2}{c} }
    0 & 0 \\ 0 & k_2
    \end{array} \right]
\end{gathered}
\end{equation}
and only one flat variable $y_1=q_1+q_2$ is required. The relationships between $y_1$ and $q_2$ and their derivatives are
\begin{equation}\begin{gathered}
    q_2 = -\frac{I_2^*}{k_2}\ddot{y}_1 + \frac{I_2^*c_2}{k_2^2}y_1^{(3)}, \quad 
    \ddot q_2 = -\frac{I_2^*}{k_2}y_1^{(4)} + \frac{I_2^*c_2}{k_2^2}y_1^{(5)}
\end{gathered}
\end{equation}
The approximated motor torque $\tau_1$ of the first joint is given by Eq.\eqref{eq:tau} and becomes
\begin{equation}\label{eq:tau2}
    \tau_{1} = I_{1}^* \ddot y_1 + \frac{I_2^*\left(I_{1}^* - I_2^*\right)}{k_2} y_1^{(4)} \\ - \frac{I_2^*\left(I_{1}^* - I_2^*\right) c_2}{k_2^2} y_1^{(5)}    
\end{equation}
The 12 boundary conditions at the beginning and the end of the motion are 
\begin{equation}
    \begin{cases}
        y_1(0)=q_{1i}+q_{2i} \\
        y_1(T)=q_{1f}+q_{2f} \\
        y_1^{(i)}(0)=0\qquad\forall\,i=1,2,\dots,5 \\
        y_1^{(i)}(T)=0\qquad\forall\,i=1,2,\dots,5
    \end{cases}
\end{equation}
where $q_{1i},q_{2i}$ and $q_{1f},q_{2f}$ are the initial and final angular positions of joints 1 and 2, respectively.
\rev{The condition $\frac{c_2}{k_2}\omega \ll 1$, used to derive Eq.~\eqref{eq:tau}, is validated by examining the frequency content of the resulting motion laws for the joint variable $q_2$. The analysis indicates a peak angular frequency of approximately 1\,Hz. Consequently, the condition $\frac{c_2}{k_2}\omega \approx 0.03 \ll 1$ is satisfied.}
\rev{It is worth noting that the numerical simulations are performed on a 2-DOF robot; however, the proposed approach readily extends to an $n$-DOF system. For instance, in \cite{Tonan20241262}, a flatness-based control strategy was applied to a 4-DOF underactuated planar robot.}

Three scenarios \rev{differing in damping values} are considered, with a \rev{point-to-point} motion having $q_{1i}=q_{2i}=0$, $q_{1f}=\pi, q_{2f}=0$, and $T=1$\,s \footnote{Please note that in a point-to-point motion $q_{2i}=q_{2f}=0$; otherwise, the robot could not hold the initial and final positions with null velocities and accelerations due to the spring placed in the second joint \cite{Bottin2022}.}.
For each scenario, the motion is planned using the polynomial law, the min-control-effort strategy with $r=150$, and the min-potential-energy strategy with $p=17$, to give the planned flat variable $y_1(t)$. 
For each motion-planning strategy, the corresponding joint 1 torque $\tau_1$ is calculated using Eq.\eqref{eq:tau2}.
Finally, the dynamic model in Eq.\eqref{Eq:sist_eq_nDOF_matrix} with the system matrices in Eq.\eqref{eq:M2C2K2} is simulated using $\tau_1$ as a feed-forward torque, in order to compare the actual motion of the robot to the planned one. The employed robot dataset is reported in Table \eqref{tab:robot}. Results are summarized in Table \eqref{tab:results} and will be commented in the following sections.
\begin{table}[t!]
\caption{2-DOF robot parameters.}
\label{tab:robot}
\begin{center}
\begin{tabular}{ccc}
\hline
\hline
Parameter & Value & Unit \\
\hline
$I_1^*$ & 6.4$\cdot10^{-4}$ & kgm$^2$ \\
$I_2^*$ & 3.3$\cdot10^{-5}$ & kgm$^2$ \\
$k_2$ & 3$\cdot10^{-3}$ & Nm/rad \\
$c_2$ & 2$\cdot10^{-5}$ & Nms/rad \\
\hline
\hline
\end{tabular}
\end{center}
\end{table}

\subsection{Scenario 1}

\newcommand{\scenariowidth}{0.95}

\begin{figure}[b!]
    \centering
    \includegraphics[width=\scenariowidth\linewidth]{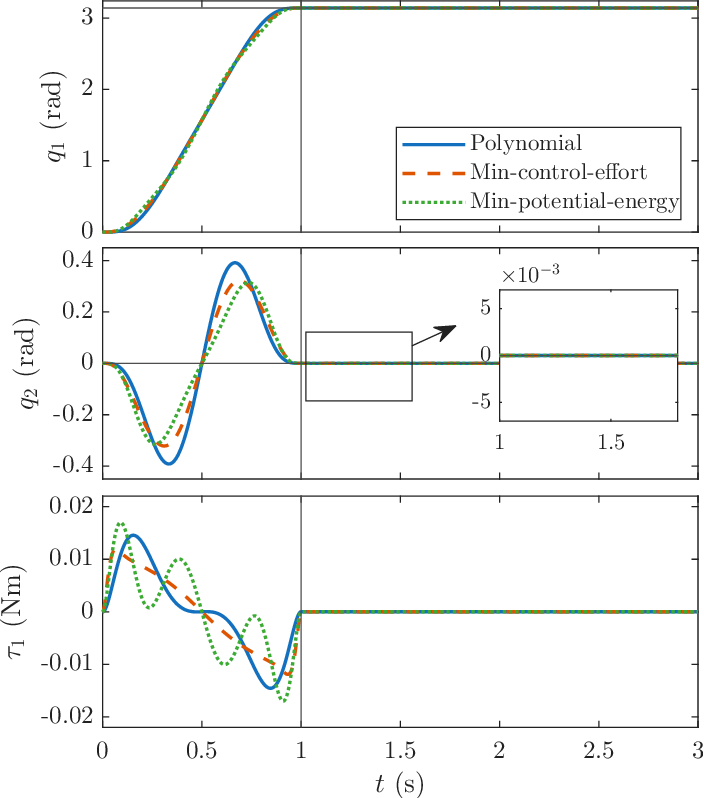}
    \caption{Angular positions of joint 1 (top) and 2 (middle) and joint 1 torque (bottom) obtained for scenario 1. The vertical solid line denotes the end of the motion.}
    \label{fig:scenario1}
\end{figure}

In the first scenario, no damping is considered (i.e. $c_2=0$), and exactly the same $c_2,k_2$ are employed in both the motion planning and simulated model. 
Figure \ref{fig:scenario1} shows the simulated $q_1,q_2$, along with the joint 1 torque $\tau_1$. The RMS of the joint 1 torque is 0.0074\,Nm for the polynomial motion planning (solid blue). Not surprisingly, this reduces to 0.0069\,Nm (-6.8\%) when using the min-control-effort strategy (dashed red). 
When the motion is planned using the min-potential-energy strategy (dotted green), the RMS of the joint 1 torque again increases to 0.0087\,Nm (+18\%), with a more oscillating control during the motion. In all cases, at the end of the motion (vertical solid line) no oscillations are observed in $q_1,q_2$, as Eq.\eqref{eq:tau2} is exact for $c_2=0$ and the same $c_2,k_2$ are employed in both the motion planning and simulated model.


\subsection{Scenario 2}

\begin{figure}[b!]
    \centering
    \includegraphics[width=\scenariowidth\linewidth]{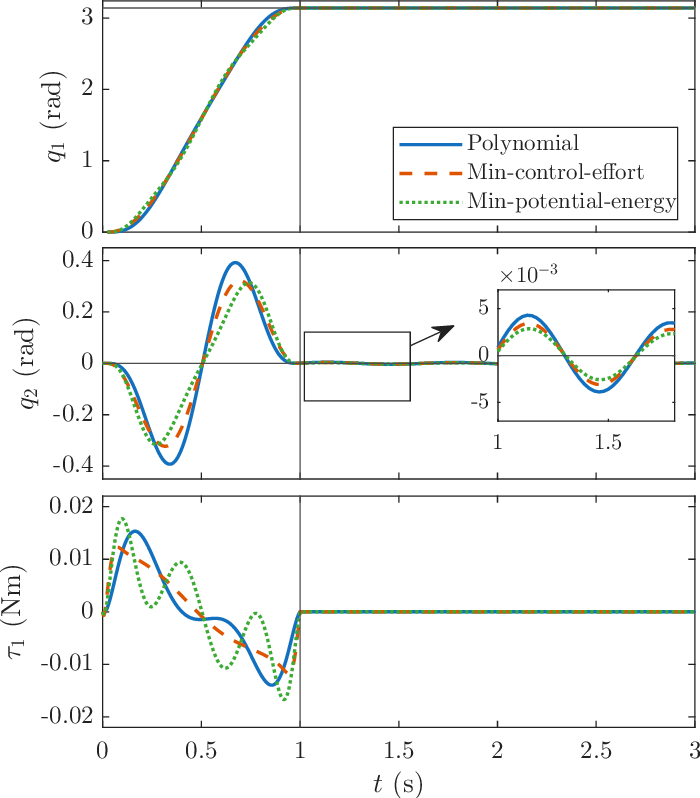}
    \caption{Angular positions of joint 1 (top) and 2 (middle) and joint 1 torque (bottom) obtained for scenario 2. The vertical solid line denotes the end of the motion.}\label{fig:scenario2}
\end{figure}

In the second scenario, damping is included (i.e. \rev{$c_2=2\cdot10^{-5}$ Nms/rad}), again with both motion planning and simulated model using the same $c_2,k_2$. The results are depicted in Figure \ref{fig:scenario2}. 

In this case, the RMS of the joint 1 torque is 0.0068\,Nm, 0.0064\,Nm (-5.9\%), and 0.0079\,Nm (+16\%) for the motion planned using the polynomial, the min-control-effort strategy, and the min-potential-energy strategy, respectively. Again, a lower RMS torque is observed with the min-control-effort strategy, whereas torque oscillations occur with the min-potential-energy strategy. 
As Eq.\eqref{eq:tau2} is approximated when $c_2\neq0$, oscillations are clearly observed in $q_2$. The initial amplitude of the oscillation in $q_2$ is 0.0045\,rad with the polynomial planning. 
The value reduces to 0.0036\,rad (-20\%) when the min-control-effort strategy is used. The reduction is even more significant when the min-potential-energy strategy is adopted, with an oscillation amplitude of 0.003\,rad (-35\%). 


\subsection{Scenario 3}

\begin{figure}[b!]
    \centering
    \includegraphics[width=\scenariowidth\linewidth]{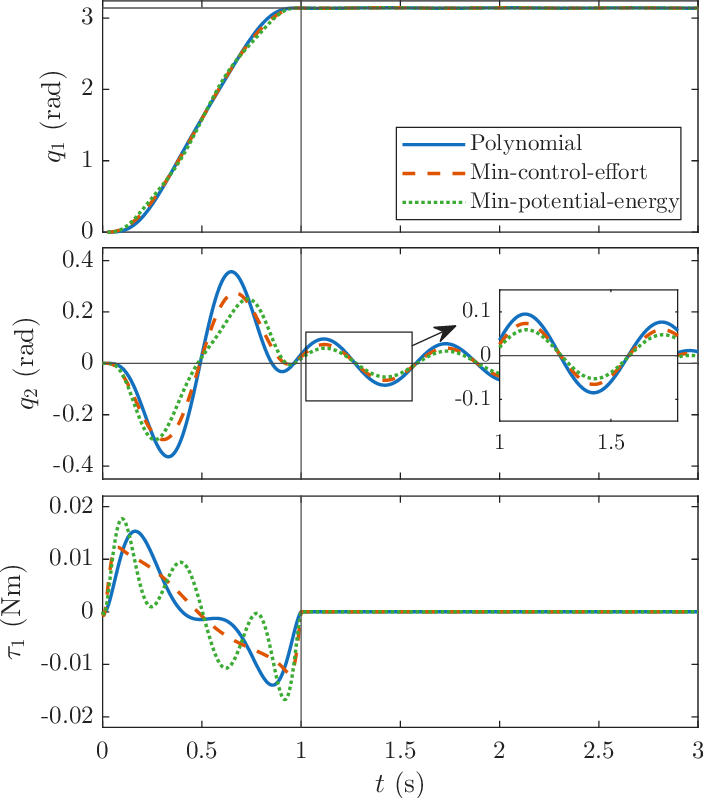}
    \caption{Angular positions of joint 1 (top) and 2 (middle) and joint 1 torque (bottom) obtained for scenario 3. The vertical solid line denotes the end of the motion.}\label{fig:scenario3}
\end{figure}
\begin{table}[tbp!]
\caption{Results for the three simulated scenarios. The percentages refer to the Polynomial strategy of the same scenario.}
\label{tab:results}
\centering
\begin{tabular}{cccc}
\hline
\hline
Scenario & Strategy & \begin{tabular}[c]{@{}c@{}c@{}}Motor \\ torque RMS \\ (Nm)\end{tabular} &
\begin{tabular}[c]{@{}c@{}}Oscillation \\amplitude at $t=T$ \\ (rad)\end{tabular} \\
\hline
 & Polynomial & 0.0074 & 0 \\
1 & Min-control & 0.0069 ($-$6.8\%) & 0 \\
 & Min-potential & 0.0087 ($+$18\%) & 0 \\\hline
  & Polynomial & 0.0068 & 0.0045 \\
2 & Min-control & 0.0064 ($-$5.9\%) & 0.0036 ($-$20\%) \\
 & Min-potential & 0.0079 ($+$16\%) & 0.003 ($-$35\%) \\\hline
  & Polynomial & 0.0068 & 0.099 \\
3 & Min-control & 0.0064 ($-$5.9\%) & 0.077 ($-$22\%) \\
 & Min-potential & 0.0079 ($+$16\%) & 0.062 ($-$38\%) \\
\hline
\hline
\end{tabular}
\end{table}


In the third scenario, a mismatch of $c_2,k_2$ between the motion planning and simulated model is considered. In particular, the damping $c_2$ and stiffness $k_2$ are both increased by 10\% in the perturbed system. The results are shown in Figure \ref{fig:scenario3}.
The joint 1 torque remains exactly the same of scenario 2, since the motion is planned with exactly the same $c_2,k_2$.
However, when the control is applied to the perturbed system, larger oscillations are clearly observed in $q_2$ due to the mismatch of $c_2,k_2$.
The amplitude of the oscillation in $q_2$ is 0.099\,rad with the polynomial planning, which is reduced to 0.077\,rad (-22\%) when the min-potential-energy strategy is employed. Again, the reduction is even more significant when using the min-potential-energy strategy, with an oscillation amplitude of 0.062\,rad (-38\%). 



\section{Conclusions}\label{sec:conclusions}

The differential flatness theory can be used to perform point-to-point motions for underactuated serial robots with specific mass distribution. 
However, in its standard form the trajectory planning is performed by means of high-degree polynomials, whereas the equations do not require such a specific planning function.
Hence, optimal control can be employed to enforce the required boundary conditions while simultaneously optimizing other factors in the trajectory, \rev{depending on the desired application}.
Results have shown that optimal control can provide better performance for differentially flat underactuated robot in terms of torque RMS (in the case of the minimal control effort planning strategy) and in terms of oscillation amplitude at the end of the motion (in the case of the minimal potential energy planning strategy). A combination of these two strategies can be employed using specific weighting factors.

\addtolength{\textheight}{-0cm}   









\bibliographystyle{ieeetr}
\bibliography{bib}

\end{document}